%% file: main.tex
\documentclass[10pt,twocolumn,letterpaper]{article}

\usepackage[pagenumbers]{cvpr} 

\usepackage{comment}
\usepackage{amsfonts,amsmath,amssymb, dsfont}
\usepackage{multirow}
\usepackage{array}
\usepackage{subcaption}
\usepackage{gensymb}
\usepackage{booktabs} 
\usepackage{float}    

\usepackage{arydshln}
\usepackage{booktabs}  
\usepackage{xcolor}    
\usepackage{colortbl}  

\usepackage[accsupp]{axessibility}  

\newcommand\blfootnote[1]{
    \begingroup
    \renewcommand\thefootnote{}\footnote{#1}
    \addtocounter{footnote}{-1}
    \endgroup
}

\input{preamble}

\definecolor{cvprblue}{rgb}{0.21,0.49,0.74}

\definecolor{lightestgray}{rgb}{0.85, 0.85, 0.85} 

\usepackage[pagebackref,breaklinks,colorlinks,allcolors=cvprblue]{hyperref}

\usepackage{xcolor}
\colorlet{linkequation}{red}
\usepackage[colorlinks]{hyperref}

\newcommand*{\SavedEqref}{}
\let\SavedEqref\eqref
\renewcommand*{\eqref}[1]{%
  \begingroup
    \hypersetup{
      linkcolor=linkequation,
      linkbordercolor=linkequation,
    }%
    \SavedEqref{#1}%
  \endgroup
}

\def\paperID{17} 
\def\confName{CVPR}
\def\confYear{2025}

\title{NeuRadar: Neural Radiance Fields for Automotive Radar Point Clouds}

\author{Mahan Rafidashti$^{\dagger, 1, 2}$
\quad
Ji Lan$^{\dagger, 1, 2}$
\quad
Maryam Fatemi$^1$
\quad
Junsheng Fu$^1$
\\
Lars Hammarstrand$^2$
\quad
Lennart Svensson$^2$
\\
\normalsize$^1$Zenseact \hspace{0.8cm} $^2$Chalmers University of Technology \hspace{0.8cm}\\
{\tt\small \{firstname.lastname\}@\{zenseact.com, chalmers.se\}}
}
\begin{document}

\input{sec/top_figure}
\input{sec/0_abstract}    
\input{sec/1_intro}

\input{sec/2_related_work}

\input{sec/2-3_preliminary}

\input{sec/3_method}

\input{sec/3-4_datasets}

\input{sec/4_experiments}

\input{sec/5_conclusion}

{
    \small
    \bibliographystyle{ieeenat_fullname}
    \bibliography{main}
}

\input{sec/X_suppl}

\end{document}

%% file: preamble.tex
\newcommand{\parsection}[1]{\noindent\textbf{#1}:}

%% file: sec/top_figure.tex
\twocolumn[{
    \maketitle
    \vspace{-12mm}
    \begin{center}
        \captionsetup{type=figure}
        \includegraphics[width=\textwidth,trim={0cm 0cm 0cm 0cm},clip]{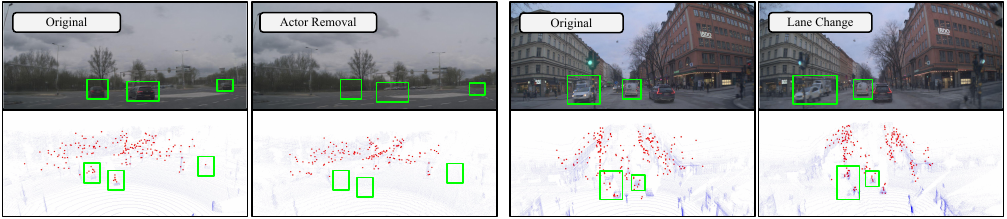}
        \captionof{figure}{NeuRadar generates radar point clouds alongside camera and lidar data for novel viewpoints and altered scenes. This figure illustrates images, radar point clouds (in red), and lidar point clouds (in blue) generated by NeuRadar for two sequences in ZOD.  }
        \label{fig:frontpage}
    \end{center}
}]

%% file: sec/0_abstract.tex
\begin{abstract}

\vspace{-3mm}

Radar is an important sensor for autonomous driving (AD) systems due to its robustness to adverse weather and different lighting conditions. Novel view synthesis using neural radiance fields (NeRFs) has recently received considerable attention in AD due to its potential to enable efficient testing and validation but remains unexplored for radar point clouds. In this paper, we present NeuRadar, a NeRF-based model that jointly generates radar point clouds, camera images, and lidar point clouds. We explore set-based object detection methods such as DETR, and propose an encoder-based solution grounded in the NeRF geometry for improved generalizability. We propose both a deterministic and a probabilistic point cloud representation to accurately model the radar behavior, with the latter being able to capture radar's stochastic behavior. We achieve realistic reconstruction results for two automotive datasets, establishing a baseline for NeRF-based radar point cloud simulation models. In addition, we release radar data for ZOD's Sequences and Drives to enable further research in this field. To encourage further development of radar NeRFs, we release the \href{https://github.com/mrafidashti/neuradar}{source code} for NeuRadar. 

\vspace{-6mm}

\blfootnote{$^{\dagger}$: These authors contributed equally to this work. }

\end{abstract}

%% file: sec/1_intro.tex
\section{Introduction}
\label{sec:introduction}

Neural radiance fields (NeRFs) may become an essential tool for testing and validating autonomous systems \cite{Yang2023ReconstructingOI, 10.1007/978-3-031-73404-5_10}. For example, 
simulating realistic data from critical and near-collision scenarios allows autonomous driving (AD) companies to test key system aspects efficiently and safely, avoiding the high costs and risks associated with studying these scenarios in real traffic. Developing NeRF models capable of representing general traffic scenarios and the sensor setups used in AD is therefore of vital importance.

Several methods construct NeRF models to generate camera and lidar data \cite{yang2023unisim, tonderski2023neurad} for AD. A common approach is to decompose the field into one component for the static background and another for each dynamic object. Recent papers have introduced neural feature fields, a type of NeRF model that outputs a feature vector for each pair of 3D point and view direction. These methods show promising results for camera and lidar data, particularly for poses close to those observed in the training data.

Many AD systems also incorporate radar sensors for increased robustness. Radars offer advantages in cases where optical detection is limited, such as low visibility and adverse weather conditions, and can serve as complementary sensors to lidar and camera in automotive perception systems \cite{Borts2024_Radar_Fields, radarpointclouds, 9127853}. Unlike lidar and camera data, radar sensors often provide a sparse point cloud for each measurement scan \cite{9968089, 8569423}. Existing NeRF models \cite{Borts2024_Radar_Fields, Huang2024_DART} cannot process such data effectively, and the task of learning a joint model to reconstruct camera, lidar, and radar point clouds remains unexplored. 

In this paper, we propose a NeRF model that jointly handles camera, lidar, and radar data for AD. Our solution is based on NeuRAD and uses a joint feature field to represent all three sensors. Given the complexity of modeling radar point clouds as a function of geometry, we take a data-driven approach to learn the mapping from the feature field to radar detections, while grounding the predictions in the NeRF geometry for improved novel view synthesis. Radar detections are known for being noisy and sparse, and even slight differences in sensor and actor poses can yield markedly different radar returns for the same scene. To address this, our model can function in both deterministic and probabilistic modes: it can generate a single predicted point cloud based on the sensor's pose, or produce a random finite set that reflects the distribution of radar detections across the scene and their return probabilities. We establish a baseline and provide suitable metrics to evaluate the performance of this method using two datasets: View-of-Delft (VoD) \cite{apalffy2022}, and a new extended version of Zenseact Open Dataset (ZOD)~\cite{alibeigizod2023}, which now includes radar point cloud data for the \textit{Sequences} and \textit{Drives} of ZOD. To sum up our contributions:

\begin{itemize}
    \item We present the first NeRF model that jointly synthesizes radar, camera, and lidar data, addressing a gap in multi-sensor 3D scene representation and enhancing the scope of unified sensor data rendering in AD applications.
    \item We introduce a data-driven radar model, employing a transformer-based network to create both deterministic and probabilistic representations of radar point clouds, thereby improving adaptability to real-world scenarios.
    \item To facilitate future exploration of multi-modal NeRFs that include radar data, we release radar point cloud data for ZOD's \textit{Sequences} and \textit{Drives}, contributing valuable resources to the research community.
    \item  We establish a baseline and evaluation metrics designed for NeRF-based radar point cloud simulation, laying the groundwork for continued research in this area.
\end{itemize}

%% file: sec/2_related_work.tex
\section{Related Work}
\label{sec:related_work}

\subsection{NeRFs for Automotive Data}


The 
task of building NeRF models for autonomous driving data has recently attracted considerable attention \cite{10.1007/978-981-99-8850-1_1, yang2023unisim, tonderski2023neurad, yang2024unipad, xie2023s, 10.1007/978-3-031-73404-5_10}. MARS \cite{10.1007/978-981-99-8850-1_1} proposes a modular framework that allows users to use various NeRF-based methods for simulating dynamic actors and static backgrounds. However, since it does not natively support lidar data and its performance relies on access to depth maps, MARS' applicability to certain datasets is limited. 
UniSim \cite{yang2023unisim}, based on Neural Feature Fields (NFF) \cite{mildenhall2021nerf, mescheder2019occupancy}, enables realistic multi-modal sensor simulation for large-scale dynamic driving scenes. NeuRAD \cite{tonderski2023neurad} extends UniSim by modeling both static and dynamic elements while integrating effective sensor modeling techniques. However, these methods, including recent work like UniCal~\cite{yang2025unical}, do not consider radar and thus fail to simulate radar detections. Our work draws inspiration from NeuRAD and expands its capabilities by incorporating radar data to simulate radar point clouds effectively. 


\subsection{Radar Simulation}

The simulation of radar data can be categorized from different perspectives \cite{magosi2022survey}. Here, we briefly describe model-based radar simulation methods, which involve modeling aspects of the environment and radar operations, and data-driven radar simulation methods, which focus on the detection results on an object or the environment \cite{magosi2022survey, 8864669}. 

\parsection{Model-based Radar Simulation} Ray-tracing methods model the propagation and interaction of radar waves by treating them as rays that travel through a 3D environment, reflecting and scattering upon encountering surfaces \cite{10041901, 10130122, 10443341, 9564521, 8008120}. These methods involve creating a detailed explicit representation of the environment and objects, limiting their scalability in complex automotive scenarios, and making them computationally intensive. In contrast, our method also passes rays through the scene, but it relies on the NeRF model to implicitly represent the scene and is purely data-driven. 

In object tracking, the radar measurement models represent reflections from multiple points on extended objects, such as vehicles, using probabilistic or geometric models. These models use structures like specific high-reflectance points and plane reflectors to create realistic radar detections \cite{1689647, 9351598, 6237574, 6237597, extendedot}. While useful for modeling radar reflections from single objects, these models include many simplifying assumptions and are unsuitable for complex environments including static and dynamic objects.

\parsection{Data-driven Radar Simulation} Data-driven models offer the flexibility to capture real-world phenomena that traditional physics-based models struggle to emulate. In \cite{10447724}, the authors simulate point clouds using a CNN to estimate the parameters of a Gaussian Mixture Model based on front camera images. However, in this method, radar detection simulation is restricted to the front camera's field of view.

Recent interest has emerged in using neural rendering to represent various types of radar data. One notable application is in Synthetic Aperture Radar (SAR), where NeRF-inspired models enhance 3D scene reconstruction~\cite{Lei2023_SAR_NeRF, Jamora2023_SAR, Liu2023_RaNeRF}. Radar Fields~\cite{Borts2024_Radar_Fields}, a modified NeRF, operates with raw frequency-domain Frequency-Modulated Continuous-Wave (FMCW) radar data, learning implicit scene geometry and reflectance models. By bypassing conventional volume rendering and directly modeling in frequency space, this approach effectively extracts dense 3D occupancy data from 2D range-azimuth maps. Another direction is explored by DART (Doppler-Aided Radar Tomography)~\cite{Huang2024_DART}, which uses radar's unique capability to measure motion through Doppler effects. DART applies a NeRF-inspired pipeline to synthesize range-Doppler images from new viewpoints, implicitly capturing the tomographic properties of radar signals. Notably, none of the aforementioned methods have directly used radar point clouds. 

\subsection{Object Detection}

The initial series of object detectors using deep learning addressed object detection indirectly by defining surrogate regression and classification problems on a substantial number of proposals~\cite{cai2019cascade, ren2016faster}, anchors~\cite{ross2017focal}, or predefined grids of potential object locations~\cite{zhou2019objects}. Alternatively, object detection can be viewed as a direct set prediction task. DETR~\cite{carion2020endtoend} employs a transformer-based model to infer a fixed number of predictions and uses a set-based global loss that forces unique matching between these predictions and the ground truth through bipartite matching. In \cite{Chen2022EfficientDO}, the authors introduce a Decoder-Free Fully Transformer-based object detection network. Following these works, we view radar detection as a set prediction problem, integrating an NFF with an encoder-only transformer to extract radar detections.

In applications requiring reliable object detection, such as autonomous driving, capturing uncertainties is crucial for safe and robust decision-making. Probabilistic object detection methods provide probability distributions for each detection, rather than fixed outputs, helping assess detection reliability and reduce overconfident errors \cite{9525313,Hall_2020_WACV, mcallister_2017, timans2024conformalod}. A probabilistic approach to DETR's set prediction problem is explored in \cite{Hess_2022}, where the predicted set of detections is modeled as a Random Finite Set (RFS). The authors use Negative Log Likelihood as a loss function and metric, which takes into account the uncertainty of the predictions \cite{Pinto_2021}. To acknowledge the stochastic nature of radar detections, we incorporate ideas from \cite{Hess_2022} into our model, enabling it to model the distribution of radar detections.  

%% file: sec/2-3_preliminary.tex
\section{Preliminaries} \label{sec:neurad}
As our method builds upon NeuRAD, we briefly describe its essential components. We explain the neural feature field, which is the core of NeuRAD. In addition, we present the modeling of camera and lidar sensors and the construction of corresponding losses. The objective is to present sufficient details to understand our radar model extension. 

\subsection{Neural Feature Field} \label{sec:nff}
At its core, NeuRAD learns a Neural Feature Field (NFF), based on iNGP 
\cite{mueller2022instant}. The NFF is defined as a continuous and learned function with learnable parameters $\theta$:
\begin{equation}\label{eq:nff}
    (s, \mathbf{f}) = f_{\theta}(\mathbf{x}, t, \mathbf{d}),
\end{equation}
where the inputs are a 3D point $\mathbf{x} \in \mathbb{R}^3$, time $t \in \mathbb{R}$, and a normalized view direction $\mathbf{d} \in \mathbb{R}^3$. 
These inputs are mapped to an implicit geometry described by the signed distance function (SDF) $s \in \mathbb{R}$ and a feature vector~$\mathbf{f} \in \mathbb{R}^{N_f}$. 

In an NFF, a ray originating from the sensor center $\mathbf{o}$ with a specific normalized view direction $\mathbf{d}$ is defined as
\begin{equation}
    \mathbf{r}(\tau) = \mathbf{o} + \tau \mathbf{d},
    \label{eq:ray_def}
\end{equation}
where $\tau$ represents the distance along the ray from the origin $\mathbf{o}$. To render the feature for the ray at time t, NeurRAD samples $N_\mathbf{r}$ 3D points $\mathbf{x}_i = \mathbf{o} + \tau_i \mathbf{d}$ and extract the corresponding feature vector $\mathbf{f}_i$ and implicit geometry $s_i$ using \eqref{eq:nff}. The final ray feature is formed using volume rendering:
\begin{equation}
    \mathbf{f}(\mathbf{r}) = \sum_{i = 1}^{N_\mathbf{r}} \omega_i \mathbf{f}_i, \,\,\,\, \omega_i = \alpha_i \prod_{j = 1}^{i - 1} (1 - \alpha_j),
    \label{eq:volrenfeature}
\end{equation}
where $\alpha_i$ represents the opacity at the point $\mathbf{x}_i$. 
This opacity is approximated by $\alpha_i = 1 / (1 + e^{\beta s_i})$, 
where $\beta$ is a learnable parameter. Furthermore, $w_i$ is the product of opacity and the cumulative transmittance to $\tau_i$ along the ray.

\subsection{Sensor Modeling} \label{sec:neurad_sensor_model}

NeuRAD models the camera by generating a feature map $\mathbf{F} \in \mathbb{R}^{H_f \times W_f \times N_f}$ by volume rendering a set of $H_f \times W_f$ camera rays, following \eqref{eq:volrenfeature}. In accordance with \cite{yang2023unisim}, this feature map is converted to an RGB image $\mathbf{I}_{\text{rgb}} \in \mathbb{R}^{H_I \times W_I \times 3}$ using a 2D CNN. In practice, the feature map is generated at a lower spatial resolution $H_f \times W_f$ compared to the final image resolution $H_I \times W_I$, drastically reducing the number of ray queries needed. 


The lidar is treated similarly, but instead of rendering an image, NueRAD renders depth, intensity, and ray drop probability for each lidar ray. In the same way as the camera, a lidar feature is rendered for each ray $\mathbf{r}(\tau) = \mathbf{o} + \tau \mathbf{d}$ of the lidar using \eqref{eq:volrenfeature}. However, in contrast to the camera model, intensity and ray drop probability are predicted independently per ray by processing each rendered ray feature through a Multi-Layer Perceptron (MLP). Further, the ray depth, which is the depth at which the ray is reflected off the scene towards the sensor, is modeled directly using the sampled 3D points as the expected depth of the ray
\begin{equation}
    D(\mathbf{r}) = \sum_{i=1}^{N_r} w_i \tau_i,
    \label{eq:depth}
\end{equation}
where $w_i$ is the same as the term used in \eqref{eq:volrenfeature}.

\subsection{Image and Lidar Losses} \label{sec:neurad_loss}

NeuRAD simultaneously optimizes all model components, employing both camera and lidar data for supervision to construct the joint loss:
\begin{equation}
    \mathcal{L}^{\text{NeuRAD}} = \mathcal{L}^{\text{image}} + \mathcal{L}^{\text{lidar}}. 
    \label{eq:neurad_loss}
\end{equation}
The image loss $\mathcal{L}^{\text{image}}$ is computed patch-wise and summed across multiple patches, and includes a reconstruction loss $\mathcal{L}^{\text{rgb}}$ and a perceptual loss $\mathcal{L}^{\text{vgg}}$. 

For the lidar loss we have four terms, one reconstruction loss for each output, \ie, depth loss $\mathcal{L}^d$ and intensity loss $\mathcal{L}^{\text{int}}$, a binary cross-entropy loss for the ray drop probability $\mathcal{L}^{p_d}$, and an additional term to suppress rendering weights of distant samples $\mathcal{L}^w$. These terms are considered independently for each lidar ray.


%% file: sec/3_method.tex
\section{Method}
\label{sec:method}

We aim to generate radar point clouds from novel viewpoints using a neural representation of a 3D scene. Since radar point clouds are sparse and noisy, we propose training our model using camera, lidar, and radar data simultaneously. This allows us to leverage the rich semantic and geometric information from camera and lidar data to inform our radar model about the environment. To learn a model of automotive scenes using all three modalities, we build upon NeuRAD and extend it to include radar data. We call this extended model NeuRadar.

\begin{figure*}[ht]
    \centering
    \includegraphics[width=1.0\linewidth]{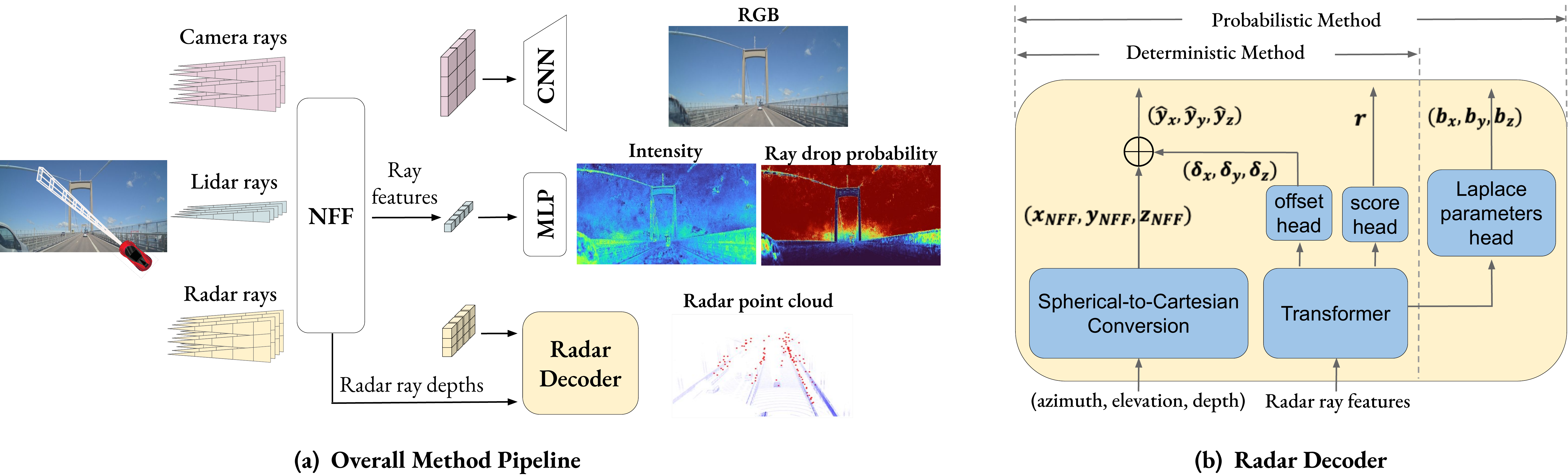}
    \caption{NeuRadar: our multimodal novel view rendering method for autonomous driving. (a) Rays from each sensor modality render ray features from the NFF. Camera and lidar branches decode their ray features into RGB values via an upsampling CNN and into lidar ray drop probability and intensity via MLPs, respectively. Radar ray features, along with estimated depths from the NFF, generate a radar point cloud via a specialized radar decoder. (b) The ray return position $({x}_{\text{NFF}}, {y}_{\text{NFF}}, {z}_{\text{NFF}})$ is obtained from the known azimuth, elevation, and estimated depth, while a Transformer predicts offsets $(\delta_x, \delta_y, \delta_z)$ and a detection confidence score $r$. The sum of the ray return position and offsets determines the final position. As an extension, the probabilistic method uses a Laplace parameter head to predict the scaling parameters. The set $\{r,\, {x}_{\text{NFF}},\, \delta_{x},\, b_{x},\, {y}_{\text{NFF}},\, \delta_{y},\, b_{y},\, {z}_{\text{NFF}},\, \delta_{z},\, b_{z}\}$ provides all necessary information for modeling the MB RFS.
    }
    \label{fig:method}
\end{figure*}

\subsection{Radar Modeling}

Although both automotive lidar and radar generate point clouds via transmitted and reflected electromagnetic waves, their measurement principles and output differ significantly. Radar’s longer wavelengths (millimeter-scale versus lidar's nanometers) make conductive materials like metal highly visible, while small objects like raindrops remain nearly undetectable. Additionally, AD radars have a larger beam width \cite{9760734}, which causes reflecting waves from close structures to interact, producing complex reflection patterns. Thus, while camera and lidar data can be modeled using local features from narrow beams in NeRF, radar detections can strongly depend on the surroundings, motivating a different modeling approach. 

Various approaches to modeling radar detections could be explored. One approach is to mimic the lidar model in NeuRAD and assume that radar detections appear at the expected depth along radar rays (see \cref{eq:depth}). However, this model does not fully account for radar properties, as radar detections can appear far from surfaces. Another approach is to formulate the task as an object detection problem, where a conventional detection network predicts detection positions directly from volume-rendered NFF features. However, this approach gives the decoder complete flexibility and requires the network to predict radar point positions entirely without using the geometry provided by the NFF. In \cref{sec:detrdecoder}, we empirically show that a naive version of this approach, which tends to memorize point clouds, is incapable of true novel view synthesis.

We address these deficiencies by combining the strengths of the two discussed approaches. Specifically, we extract the expected depth from the NFF, leveraging NeRF geometry to enhance novel view synthesis. Meanwhile, we use a simple transformer to model the difference between where the simulated radar rays bounce (estimated using \eqref{eq:depth}) and the positions of radar detections, effectively accounting for radar properties. Our method is illustrated in \cref{fig:method} and described in detail in the following subsections. 




\subsubsection{Radar Feature Rendering} \label{sec:radar_feature_rendering}

Given the position and pose of the radar sensor in space, our objective is to predict its output: a set of points in 3D space. We estimate the spatial coordinates of these points along with confidence measures that indicate the likelihood that each predicted point corresponds to an actual radar reflection. In addition, we quantify the uncertainty in spatial coordinates for each radar point. To this end, we extract relevant information from the NFF--specifically, the expected depth and a feature vector for each point in the scene--and use a decoder network to predict the point cloud coordinates and other parameters. 
 
\begin{figure}[ht]
    \centering
\includegraphics[width=0.7\linewidth]
{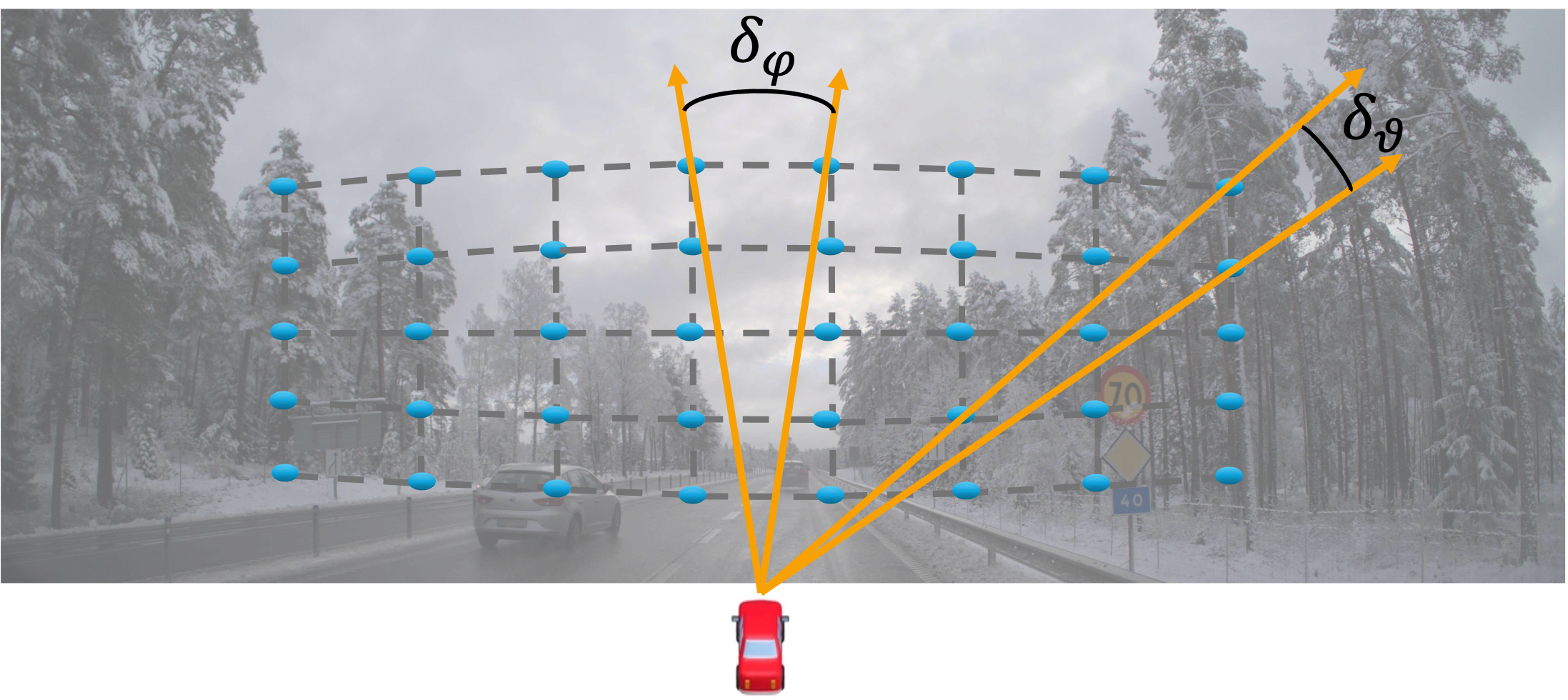}
    \caption{Illustration of our design, where radar-projected rays cover the full sensor field of view (FOV). The virtual grid (grey dashed lines) indicates the complete FOV, with blue points denoting uniformly sampled locations. The direction from the sensor to each blue point defines a ray. 
    }
    \label{fig:radar_ray_generation}
\end{figure}

A key step in this process is constructing a set of rays that traverse the scene, extracting spatial and feature information from the NFF for our radar decoder model. To retrieve scene information, we parameterize these rays as 
\begin{align}\label{eq:radar_rays}
    \{\mathbf{r}_{jk}(\tau): \mathbf{r}_{jk}(\tau) = \mathbf{o} + \tau\mathbf{d}_{jk}\}_{j,k = 1}^{N_\varphi, N_\vartheta},
\end{align}
where $\mathbf{o}$ is the sensor origin and the ray directions $\mathbf{d}_{jk}$ are unit vectors parameterized by azimuth angle $\varphi_j$ and elevation angle $\vartheta_k$ which are uniformly spaced within the radar field of view. The number of rays in azimuth ($N_\varphi$) and elevation ($N_\vartheta$) is determined by the radar field of view and ray divergence hyperparameters ($\delta_\varphi$ and $\delta_\vartheta$). \cref{fig:radar_ray_generation} illustrates this design, which ensures that radar ray rendering captures all detectable scene information.

For each ray in the set, we compute a feature vector and the expected depth of the reflected ray using volumetric rendering, as described in \eqref{eq:volrenfeature} and \eqref{eq:depth}, respectively. When generating a single ray, the azimuth and elevation information is known. With the supplementary information of the expected depth from the NFF, we determine the ray return positions, consisting of $(azimuth, elevation, depth)$. This spherical coordinate is then transformed into Cartesian coordinates and embedded into a feature vector. The final feature vector for a single ray is then obtained by adding the embeddings to the original feature vector obtained from \eqref{eq:volrenfeature}. The final feature vector, hence, combines features from the NFF with knowledge about the geometry. As our radar rays share the NFF with the lidar and camera model, the model leverages information about the 3D geometry of the scene obtained from these sensors when rendering its features.



\subsubsection{Radar Decoder}
\label{sec:radarmodeling}

The radar decoder takes the extracted features and predicts the radar point cloud. The overall architecture contains a transformer encoder that performs self-attention across the radar ray features, followed by small MLP heads for the different prediction tasks, see \cref{fig:method} (b). 

We formulate one deterministic and one probabilistic version of the decoder. 
The deterministic method predicts offsets from the known ray return positions. By adding the offsets to the known positions, we obtain the desired set of 3D points. The decoder also predicts a confidence score for each point representing the probability that it corresponds to an actual detection. By thresholding these scores, the model outputs the estimated radar point cloud. As an extension, the probabilistic method predicts additional parameters to model the distribution of the radar point cloud. In this way, the output is a distribution rather than a fixed set of points, allowing different radar point clouds to be sampled. 

\subsubsection{Point Cloud Representation}\label{sec:pointcloudrep}

We represent the radar point cloud using both deterministic and probabilistic approaches. In the deterministic approach, for each of the rays, we obtain the Cartesian position of the ray return, $\hat{\mathbf{y}}_{\text{NFF}} = (x_{\text{NFF}}, y_{\text{NFF}}, z_{\text{NFF}})^T$, from the NFF. We then predict an offset from these coordinates, $\delta_{\hat{\mathbf{y}}} = (\delta_x, \delta_y, \delta_z)^T$, and a detection confidence score $r$, using the transformer. The final detection position, $\hat{\mathbf{y}} = (\hat{\mathbf{y}}_x, \hat{\mathbf{y}}_y, \hat{\mathbf{y}}_z)^T$, is computed by $\hat{\mathbf{y}} = \hat{\mathbf{y}}_{\text{NFF}} + \delta_{\hat{\mathbf{y}}}$.

With this approach, the model directly renders a point cloud that is deterministic in the sense that, for a given input, the positions of the detections are fixed. From a simulation perspective, this could be desirable, \eg, to ensure repeatability when testing downstream tasks.

\begin{figure}
    \centering
\includegraphics[width=1.0\linewidth]{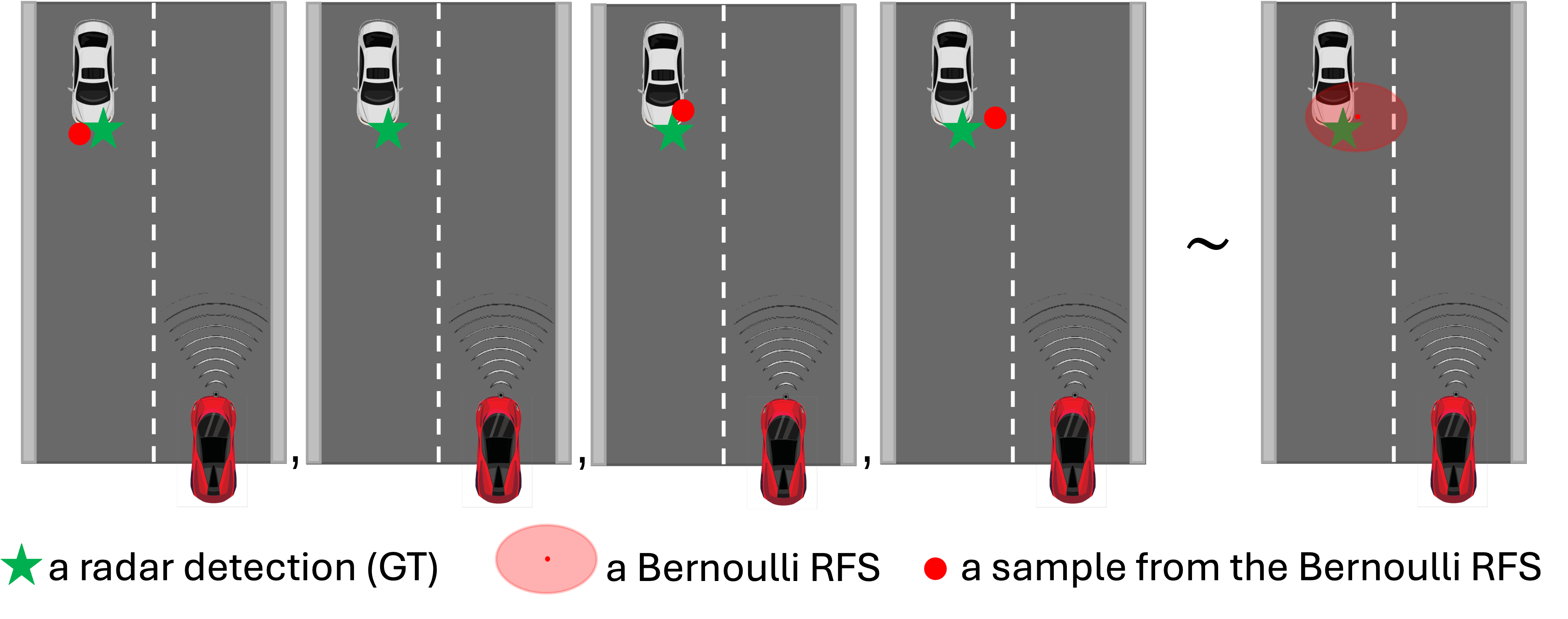}
    \caption{Four sampled sets (left) drawn from a Bernoulli RFS (right) with an existence probability of $r=0.75$. The RFS accounts for both the absence of objects and spatial uncertainties. The image is provided for context only.}
    \label{fig:bernoulli_rfs}
\end{figure}

We also provide a probabilistic representation of the radar point cloud to capture uncertainties in radar detections and to allow for rendering slightly different radar point clouds for a given input. Inspired by \cite{Hess_2022}, we model the probabilistic point cloud as a Multi-Bernoulli Random Finite Set (MB-RFS), 
\begin{align}\label{eq:mbrfs}
    \mathbb{Y} = \bigcup_{i=1}^{N_{\text{rays}}} \mathbb{Y}_i,
\end{align}
where $\mathbb{Y}_1,\dots,\mathbb{Y}_{N_{\text{rays}}}$ are Bernoulli RFSs describing possible individual radar detections. The radar point cloud is, thus, modeled as a set of detections for which the cardinality and the points themselves are random variables.

Each Bernoulli RFS $\mathbb{Y}_i$ is defined by
\begin{equation}
    f_{\text{B}_i}(\mathbb{Y}_i) = \begin{cases}
        1-r_i, & \text{if } \mathbb{Y}_i=\emptyset, \\
        r_ip_i(\mathbf{y}), & \text{if } \mathbb{Y}_i=\{\mathbf{y} \}, \\
        0, & \text{otherwise,}
    \end{cases}
    \label{eq:single_bernoulli}
\end{equation}
where $r_i$ is the existence probability and the density $p_i(\mathbf{y})$ describes the position of the radar detection. In practice, the existence probability $r_i$ is the same as the detection confidence score defined in the deterministic method. We also assume that the density $p_i(\mathbf{y})$ can be factorized as
\begin{equation}
     p_i(\mathbf{y}) = p_i(x)p_i(y)p_i(z) \,,
     \label{eq:posdensity}
\end{equation}
where the individual densities are modeled as Laplacian distributions parameterized by $(\mu_{x_i}, b_{x_i})$, $(\mu_{y_i}, b_{y_i})$ and $(\mu_{z_i}, b_{z_i})$, respectively. In practice, $(\mu_{x_i}, \mu_{y_i}, \mu_{z_i})$ is the same as the detection positions $(\delta_x, \delta_y, \delta_z)$ defined in the deterministic method. Thus, for each ray $i$, we predict the set of parameters $\{r_i,\, {x}_{\text{NFF}_i},\, \delta_{x_i},\, b_{x_i},\, {y}_{\text{NFF}_i},\, \delta_{y_i},\, b_{y_i},\, {z}_{\text{NFF}_i},\, \delta_{z_i},\, b_{z_i}\}$ that describe the corresponding Bernoulli distribution. 


Using \eqref{eq:mbrfs} and \eqref{eq:single_bernoulli} we can express the set probability distribution for the complete radar point cloud as a multi-Bernoulli (MB) distribution, defined as
%
\begin{equation}
    f_{\text{MB}}(\mathbb{Y}) = \sum_{\uplus_{i\in \mathbb{I}} \mathbb{Y}_i = \mathbb{Y}} \prod_{i=1}^{N_\mathbf{q}} f_{\text{B}_i}(\mathbb{Y}_i),
    \label{eq:mbdist}
\end{equation}
where $\uplus_{i\in \mathbb{I}} \mathbb{Y}_i = \mathbb{Y}$ represents the sum over all disjoint sets with union $\mathbb{Y}$. Note that each Bernoulli in the MB is independent making it convenient to generate a radar point cloud from \eqref{eq:mbdist} by generating samples from each Bernoulli separately using \eqref{eq:single_bernoulli}. An example of a Bernoulli RFS prediction and corresponding samples is shown in \cref{fig:bernoulli_rfs}.

\subsection{Losses}
\label{sec:losses}

Our total loss is formed by extending the NeuRAD loss defined in \eqref{eq:neurad_loss} with a radar loss $\mathcal{L}^{\text{radar}}$,
\begin{equation}
    \mathcal{L} = \mathcal{L}^{\text{NeuRAD}} + \mathcal{L}^{\text{radar}}.
\end{equation}
As previously mentioned, we consider two types of radar models, one deterministic and one probabilistic, each having its own formulation of $\mathcal{L}^{\text{radar}}$. In this section, we present both of these losses and how we match ground truth radar detections to model predictions to calculate the loss.

\parsection{Matching}
\begin{subequations}
Similar to \cite{Hess_2022, carion2020endtoend}, we find a matching or an assignment between predictions and ground truth by solving an optimal assignment problem. Let $\mathbb{Y} = \{\mathbf{y}_j\}_{j=1}^{|\mathbb{Y}|}$ denote the ground truth set of radar points, and $\hat{\mathbb{Y}} = \{\hat{\mathbf{y}}_i\}_{i=1}^{|\hat{\mathbb{Y}}|}$ the set of radar point predictions; for the probabilistic model, we set $\hat{\mathbf{y}}_i=(\mu_{x_i},\mu_{y_i},\mu_{z_i})^\text{T}$. Note that $r_i$ represents the probability that $\hat{\mathbf{y}}_i$ is a reflected radar point. We ensure that $|\hat{\mathbb{Y}}| > |\mathbb{Y}|$ by setting the number of rays such that it is larger than the maximum number of radar detections across all scans in the dataset. The assignment problem can be formulated as
    \renewcommand{\theequation}{\theparentequation\alph{equation}}
    \begin{align}
        \min_{\mathbf{A}} \quad & \sum_{i} \sum_{j} \mathbf{C}_{i,j} \mathbf{A}_{i,j}, \label{eq:cost_objective} \\
        \text{s.t.} \quad & \sum_{i=1}^{|\hat{\mathbb{Y}}|} \mathbf{A}_{i,j} = 1,  \quad \sum_{j=1}^{|\mathbb{Y}|} \mathbf{A}_{i,j} \leq 1, \label{eq:assign_constraint} \\
        & \mathbf{C}_{i,j} = ||\hat{\mathbf{y}}_i-\mathbf{y}_j|| -\log \left(r_i\right), \label{eq:cost}
    \end{align}
\end{subequations}
where $\mathbf{C}$ is a cost matrix and $\mathbf{A}$ is the assignment matrix. In \eqref{eq:cost}, both the Euclidean distance between $\hat{\mathbf{y}}_k$ and $\mathbf{y}_l$, and the existence probability of the $k$th prediction, are considered jointly. Constrained by \eqref{eq:assign_constraint}, all ground truth objects will be assigned to a prediction each, while some predictions will remain unassigned. By optimizing \eqref{eq:cost_objective}, we obtain the optimal assignment, denoted by $\mathbf{A}^*$, and the set $\gamma = \{i \in \mathbb{N}_{|\hat{\mathbb{Y}}|}, j \in \mathbb{N}_{|\mathbb{Y}|} \mid \forall i, j: \mathbf{A}^*_{i,j}=1\}$, which records all matched pairs. 

\parsection{Deterministic Radar Loss}
For the deterministic radar model, the loss is defined as
\begin{equation}
    \begin{split}
        \mathcal{L}_{\text{det}}^{\text{radar}} &= \sum_{(i,j) \in \gamma} \big( ||\hat{\mathbf{y}}_i - \mathbf{y}_j|| - \log(r_i) \big)-\sum_{(i,j) \notin \gamma} \log(1-r_i).
    \end{split}
    \label{eq:radar_loss_det}
\end{equation}
In \eqref{eq:radar_loss_det}, the first term minimizes the existence probability of the unassigned predictions. Additionally, the second term reduces the Euclidean distance between matched pairs of predicted and ground truth radar points and maximizes the existence probability of the assigned predictions.

\parsection{Probabilistic Radar Loss} In the probabilistic method, we approximate the negative log-likelihood of the MB distribution $-\log f_{\text{MB}}(\mathbb{Y})$ in \eqref{eq:mbdist} by the term corresponding to $\mathbf{A}^*$ and use that as our loss:
\begin{equation}
\begin{split}
    \mathcal{L}_{\text{prob}}^{\text{radar}} 
    &= - \sum_{(i,j) \notin \gamma} \log(1 - r_i) \\
    &- \sum_{(i,j) \in \gamma} \log(r_ip_{i,x}(\mathbf{y}_{j,x})p_{i,y}(\mathbf{y}_{j,y})p_{i,z}(\mathbf{y}_{j,z})).
\end{split}
\label{eq:radar_loss_prob}
\end{equation}
Similar to $\mathcal{L}_{\text{det}}^{\text{radar}}$, $\mathcal{L}_{\text{prob}}^{\text{radar}}$ minimizes the existence probability of the unassigned predictions and maximizes the existence probability of the assigned predictions. In contrast, the second term in \eqref{eq:radar_loss_prob} maximizes the probability that a matched radar point prediction accurately represents its corresponding ground truth radar point. The similarity between $\mathcal{L}_{\text{det}}^{\text{radar}}$ and  $\mathcal{L}_{\text{prob}}^{\text{radar}}$ becomes even more striking by noting that $ \log  p_{i,x}(\mathbf{y}_{j,x}) = |\mathbf{y}_{j,x} - \mu_{x_i}|/(2b_{x_i}) + \log (2b_{x_i})$.

%% file: sec/3-4_datasets.tex
\section{Datasets}\label{sec:datasets}
In this section we provide an overview of available datasets suitable for studying automotive radar point cloud generation based on neural representations of the scene. We also introduce a new dataset by extending the existing ZOD \cite{alibeigizod2023} with radar point cloud data.

\subsection{Existing Datasets} 
To facilitate studies like the one presented in this paper, researchers need datasets that include driving sequences with good-quality camera images, lidar, and radar point clouds, along with a permissive license for broad accessibility.
Among existing AD datasets, Pandaset \cite{xiao2021pandaset}, Argoverse \cite{wilson2021argoverse2} and Kitti-360 \cite{Liao2021KITTI360AN} offer good-quality camera and lidar data with permissive licensing but lack radar point cloud data. Both nuScenes \cite{nuscenes2019} and VoD \cite{apalffy2022} provide camera images, lidar and radar point clouds with relatively permissive licenses. However, nuScenes radar point cloud is sparse, making it less suitable for the task at hand. Despite VoD's limitations in image quality, we selected it as one of the datasets used in this paper due to its dense radar point clouds. Nonetheless, we emphasize the need for automotive datasets that better meet essential requirements, including high-quality camera, lidar, and radar data, along with a permissive license to advance research.

\subsection{Zenseact Open Dataset} 
ZOD \cite{alibeigizod2023} is a diverse and large-scale multimodal dataset for autonomous driving collected in several European countries which includes a million curated \textit{Frames}, 1473
\textit{Sequences} of 20-second length and 29 extended \textit{Drives} spanning several minutes. Like many contemporary datasets, ZOD includes high-resolution lidar and camera data but was initially provided without radar data. 

In this paper, we extend ZOD \textit{Sequences} and \textit{Drives} by adding radar point cloud data. Radar point clouds are obtained from a Continental Radar ARS513 B1 sample – year 2022, and are captured at every 60 ms. 
The addition of radar data together with ZOD's permissive license enables diverse research applications in automotive perception, such as multimodal scene reconstruction and novel view synthesis, that \emph{includes} radar point cloud data. Further details on the radar data are in the supplementary material.

%% file: sec/4_experiments.tex
\section{Experiments}
\label{sec:experiments}

In this section, we evaluate NeuRadar’s ability to reconstruct radar point clouds alongside camera and lidar data for VoD and ZOD datasets. 
We evaluate the performance of the model in terms of reconstruction accuracy, and extrapolation to confirm the consistency and generalization of NeuRadar in various scenarios. 
In this work, we use Chamfer Distance (CD) \cite{barrow1977parametric} and Earth Mover's Distance (EMD) \cite{Rubner2000TheEM} to measure the similarity between the generated point cloud and the ground truth point cloud. 

As a baseline, we adapt NeuRAD's lidar point cloud generation method to generate radar detections. For each radar ray in \eqref{eq:radar_rays}, we predict the depth according to \eqref{eq:depth}, which together with ray direction gives us possible radar detection~$\hat{\mathbf{y}}$.
The probability of generating this detection is modelled using a simple network fed with the ray features \eqref{eq:volrenfeature} as input. Following the Lidar ray drop probability and intensity prediction methods, we use an MLP. 
The deterministic loss described in \eqref{eq:radar_loss_det} is then used to train this baseline model. 
The predicted existence probabilities are then used to decide which rays would reflect and result in radar detections. In the baseline method and the deterministic method, we choose detections with existence probability higher than a threshold, here 0.5. In the probabilistic method, we simply sample from the MB RFS. 

\subsection{Implementation Details}

NeuRadar is implemented in nerfstudio \cite{nerfstudio} and is based on the open-source NeuRAD code. For the radar decoder, the learning rate varies from $10^{-3}$ to $10^{-7}$ over the first 10,000 iterations, with a warmup period of 5,000 iterations. For the rest of the network, we use the same training schedule as NeuRAD. We train our model for 20,000 iterations, which takes on average three hours on a single NVIDIA A100.

\subsection{Novel View Synthesis}\label{sec:novelviewsynth}

\begin{figure*}[ht]
    \centering
    \includegraphics[width=0.9\linewidth]{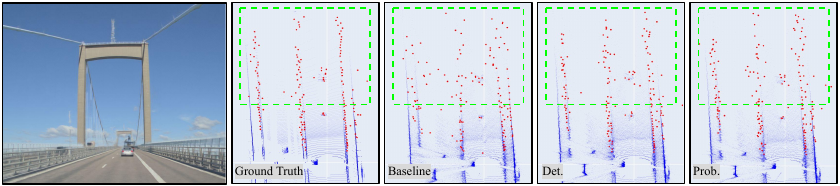}
    \caption{Novel view synthesis for interpolated frames including radar (red) and lidar (blue) detections. The Deterministic (Det.) and Probabilistic (Prob.) methods outperform the baseline at larger distances.}
    \label{fig:interzod}
\end{figure*}

To evaluate the novel view synthesis performance of NeuRadar, during training, we hold out every second frame in the sequence and test on this data. For each dataset, we train models on 10 sequences (described in \cref{sec:suppmatdata}) and report the median value of each metric over these sequences. Alongside the metrics for the entire point cloud, we report metrics for radar points within the 30 and 80-meter range of the sensor. \Cref{tab:results_det_prob} shows that both of our proposed methods outperform the baseline for both datasets, especially in higher range. The baseline method performs reasonably well near the sensor but degrades as the distance increases. This is demonstrated in \cref{fig:interzod}, which shows the radar and lidar point clouds rendered alongside the image using the three methods. Furthermore, the probabilistic NeuRadar model performs better than the deterministic model and the baseline for both VoD and ZOD. 

\begin{table*}[ht]
    \centering
    \caption{Novel view synthesis performance comparison of the baseline model, our deterministic model (Det.), and our probabilistic model (Prob.). We evaluated the results for predictions and ground truth within 30 and 80 meters of the radar sensor (indicated by the subscript) and the entire point cloud (no subscript). Lower CD and EMD values indicate better performance.}
    \resizebox{0.9\linewidth}{!}{
    \begin{tabular}{l c c c c c c c c c c c c c}
    \toprule
    \multirow{2}{*}{Method} & \multicolumn{6}{c|}{ZOD} & \multicolumn{6}{c|}{VoD}  \\
    \cmidrule(lr){2-7} \cmidrule(lr){8-13}
    & CD$_{30}$ & EMD$_{30}$  & CD$_{80}$ & EMD$_{80}$ & CD & EMD & CD$_{30}$ & EMD$_{30}$ & CD$_{80}$ & EMD$_{80}$  & CD & EMD\\ \midrule
    
    Baseline  & 2.854 & 2.981 & 4.257 & 4.460 & 5.536 & 5.851 & 2.965 & 2.799 & 3.924 & 4.067 & 4.698 & 4.966 \\
    Det.      & 2.592 & 2.647 & 3.981 & 4.050 & 4.439 & \textbf{5.136} & 2.122 & 2.405 & 3.606 & 3.256 & 3.785 & 3.890 \\
    Prob.     & \textbf{2.309} & \textbf{2.601} & \textbf{3.846} & \textbf{4.025} & \textbf{4.298} & 5.348 & \textbf{2.091} & \textbf{2.272} & \textbf{3.239} & \textbf{3.189} & \textbf{3.388} & \textbf{3.734} \\
    \bottomrule
    \end{tabular}
    }
    \label{tab:results_det_prob}
\end{table*}

\subsection{Novel Scenario Generation}\label{sec:novelscenariogen}

In this section, we explore NeuRadar’s capability to synthesize radar point clouds from entirely novel viewpoints, from which the radar has not previously observed the scene. There is no ground truth available for these viewpoints, therefore, we present qualitative results through visualizations to present NeuRadar’s ability to generalize beyond observed viewpoints. \Cref{fig:frontpage} illustrates the results of removing dynamic actors from the scene and shifting the ego pose 3 meters laterally to simulate lane change. 
The visualizations indicate that NeuRadar successfully generates a realistic radar point cloud, reinforcing its potential for novel scenario generation in autonomous driving applications.

\subsection{Ablations} \label{sec:ablations}



We present experiments on two key model design choices, conducted on our probabilistic radar model across three sequences in two datasets. The first experiment investigates two popular choices for the spatial density used in the Bernoulli distribution: Laplace and Gaussian. As shown in \cref{tab:ab_study_pdf}, models using the Laplace distribution outperform those using the Gaussian distribution. Thus, we advocate using the Laplace distribution in our probabilistic radar model.

\begin{table}[ht]
    \centering
    \caption{Results for two spatial density functions used in the MB.} 
    \resizebox{0.65\linewidth}{!}{
    \begin{tabular}{l c c c }
    \toprule
    Dataset & Density & CD $\downarrow$ & EMD $\downarrow$  \\ \midrule
    
    \multirow{2}{*}{ZOD} 
    & Laplace   & \textbf{3.92}     & \textbf{4.94} \\
    & Gaussian   & 4.91 & 5.86 \\
    \midrule

    \multirow{2}{*}{VoD} 
    & Laplace  & \textbf{4.02}  &  \textbf{4.28}  \\
    & Gaussian  & 4.28 &  4.44  \\
    
    \bottomrule
    \end{tabular}
    }
    \label{tab:ab_study_pdf}
\end{table}

The second experiment evaluates three designs in the radar decoder. As shown in \cref{tab:ab_study_decoder}, models employing a Transformer-based architecture outperform those using either an MLP or a CNN-based architecture. This is likely because the Transformer's self-attention mechanism enables the radar decoder to capture long-range dependencies and contextual information in radar data more effectively.

\begin{table}[ht]
    \centering
    \caption{Results for different models used in the radar decoder.} 
    \resizebox{0.8\linewidth}{!}{
    \begin{tabular}{l c c c c}
    \toprule
     \multirow{2}{*}{Model} & \multicolumn{2}{c}{ZOD} & \multicolumn{2}{c}{VoD} \\  \cmidrule(lr){2-3} \cmidrule(lr){4-5}
      & CD $\downarrow$ & EMD $\downarrow$ & CD $\downarrow$ & EMD $\downarrow$ \\ 
     \midrule
     MLP & 4.93 & 6.14 & 4.30 & 4.61 \\
     CNN & 5.68 & 6.29 & 4.51 & 4.97 \\
     Transformer & \textbf{3.92} & \textbf{4.94} & \textbf{4.02}  &  \textbf{4.28} \\
    \bottomrule
    \end{tabular}
    }
     \vspace{-2.1mm}
    \label{tab:ab_study_decoder}
\end{table}

%% file: sec/5_conclusion.tex
\section{Conclusion}
\label{sec:conclusion}

 In this paper, we introduced NeuRadar, a multi-modal NeRF-based simulation method that includes a data-driven radar model. NeuRadar leverages a unified neural feature field to effectively synthesize camera, lidar, and radar data and is capable of generating realistic radar point clouds from novel views and for different scene configurations. To address the challenge of noisy and sparse radar detections, we developed a probabilistic version of NeuRadar that incorporates the multi-Bernoulli distribution. This version produces a random finite set reflecting the distribution of radar detections across the scene and their return probabilities. Additionally, we enhanced the ZOD dataset by adding radar data to its \textit{Sequences} and \textit{Drives}. We established a baseline and evaluation metrics for NeRF-based radar point cloud simulation and conducted evaluations on three public AD datasets. Lastly, we release our code to foster further research into automotive radar data simulation. 

\section*{Acknowledgment}

This work was supported by Vinnova, Sweden's Innovation Agency. Computational resources were provided by NAISS at NSC Berzelius, partially funded by the Swedish Research Council, grant agreement no. 2022-06725. 

%% file: sec/X_suppl.tex
\clearpage
\setcounter{page}{1}
\maketitlesupplementary

\appendix
\renewcommand{\thesection}{\Alph{section}} 

In the supplementary material, we present details of NeuRadar's implementation, more information about ZOD's radar data, and additional visualizations. In \cref{sec:implementationdetails}, we provide information about the training process, network architecture, and losses. In addition, we provide our hyperparameter values. In \cref{sec:suppmatdata}, we present more detailed information about the sequences used to evaluate our network and ZOD's radar point cloud data. In \cref{sec:imagelidareffects}, we investigate the effects of adding radar point cloud data on image and lidar reconstruction by comparing our method to NeuRAD, followed by an exploration of using a DETR-like object detection network as a radar decoder. We then provide more figures depicting our results in \cref{sec:visualizations}. Finally, we identify the limitations of our method and outline potential directions for future work in \cref{sec:limits}. 

\section{Implementation Details}
\label{sec:implementationdetails}



\parsection{Training} We train all parts of our model jointly with 20,000 iterations, using the Adam optimizer. Regarding the number of rays in each iteration, we follow the settings in NeuRAD \cite{tonderski2023neurad} for camera and lidar, \ie, 16,384 lidar rays and 40,960 camera rays. For radar, the number of rays in each iteration is not fixed. Instead, the number of radar rays in each iteration equals the number of radar rays in each scan multiplied by the number of radar scans loaded in each iteration. The radar specifications for the ZOD and VoD are shown in \cref{tab:spec_radar_rays}. The number of rays per iteration is 54,784 for ZOD and 70,400 for and VoD.

For the optimization, we adopt the same settings for existing modules in NeuRAD. For the new module, the radar decoder, we use a warmup of 5,000 steps and a learning rate of $0.001$ that decays by an order of magnitude throughout the training. 

\begin{table*}[tb]
    \centering
    \caption{Specifications for the radar in two datasets. The unit for ray divergence is radians.}
    \vspace{-2mm}
    \resizebox{.9\linewidth}{!}{
    \begin{tabular}{l c c c c c c}
    \toprule
     Dataset & Azimuth range & Elevation range & Ray divergence & \#rays per scan & \#scans & \#rays per iteration \\ \midrule
    
    ZOD & $\pm45.84^{\degree}$  & $(-4.58^{\degree}, 22.92^{\degree})$  & $0.015 (0.8594^{\degree})$ & $3424$ & $16$  &  $54784$ \\ 
    \midrule

    VoD & $\pm57.29^{\degree}$  & $(-22.34^{\degree}, 28.07^{\degree})$  & $0.02(1.14^{\degree})$ & $4400$ & $16$  &  $70400$ \\ 
    \bottomrule
    \end{tabular}
    }
    \label{tab:spec_radar_rays}
\end{table*}

\parsection{Neural Feature Field} We use NeuRAD's hyperparameter settings for the neural feature field (NFF) in NeuRadar. Additionally, radar rays have specific hyperparameters, such as ray divergence and a scaling parameter, which are explained in \cref{sec:radar_feature_rendering}. All NFF-related settings are listed in \cref{tab:hyperparameters}.

\parsection{Hashgrids} We follow NeuRAD and employ the efficient hashgrid implementation provided by tiny-cuda-nn \cite{tiny-cuda-nn}, configuring two distinct hashgrids, one for the static environments and one for dynamic actors. For the static environments, a significantly larger hash table is employed, given that actors comprise only a minimal area of the overall scene. 


\parsection{Losses} In \cref{sec:neurad_loss} and \cref{sec:losses}, we explain that the total loss in NeuRadar comprises the radar loss $\mathcal{L}^{\text{radar}}$, image loss $\mathcal{L}^{\text{image}}$, and lidar loss $\mathcal{L}^{\text{lidar}}$, with the latter two forming the NeuRAD loss. While the paper focuses on the theory and motivation behind these losses, we present their detailed equations here. 

The image loss is computed by
\begin{equation}
    \mathcal{L}^{\text{image}} = \frac{1}{N_p} \sum_{i=1}^{N_p} (\lambda^{\text{rgb}} \mathcal{L}^{\text{rgb}}_i + \lambda^{\text{vgg}} \mathcal{L}^{\text{vgg}}_i),
\end{equation}
where $N_p$ denotes the number of patches, and $\lambda^{\text{rgb}}$ and $\lambda^{\text{vgg}}$ are weighting hyperparameters. The lidar loss is defined as
\begin{equation}
    \mathcal{L}^{\text{lidar}} = \frac{1}{N} \sum_{i=1}^{N} (\lambda^d \mathcal{L}_i^d + \lambda^{\text{int}} \mathcal{L}^{\text{int}}_i + \lambda^{p_d} \mathcal{L}_i^{p_d} + \lambda^w \mathcal{L}_i^w), 
\end{equation}
where $N$ denotes the number of lidar points, and $\lambda^d$, $\lambda^{\text{int}}$, $\lambda^{p_d}$, and $\lambda^w$ are weighting hyperparameters. Finally, the radar loss is calculated by
\begin{equation}
    \mathcal{L}^{\text{radar}} = \begin{cases}
        \lambda^{\text{radar}} \mathcal{L}_{\text{det}}^{\text{radar}}, & \text{if deterministic modeling}, \\
        \lambda^{\text{radar}} \mathcal{L}_{\text{prob}}^{\text{radar}}, & \text{if probabilistic modeling},
    \end{cases}
    \label{eq:radar_loss}
\end{equation}
where $\lambda^{\text{radar}}$ is a weighting hyperparameter. The values of these hyperparameters are given in \cref{tab:hyperparameters}. 

\parsection{Offset Head} A critical hyperparameter for offset prediction is the maximum offset value, which acts as a constraint on the radar decoder's offset head. With respect to this hyperparameter, a small value makes the estimations from NFF dominant, while a large value gives the radar decoder more flexibility. \cref{tab:ab_study_offset} presents the performance for varying maximum offset values. The experiments are conducted on our probabilistic radar model across three evaluation sequences from two datasets. For ZOD, a maximum offset of 1.5 meters yields good results in terms of both CD and EMD. However, the best CD and EMD values for VoD do not align. Averaging the metrics suggests that 1.5 meters is also suitable for VoD.


%
\begin{table}[tb]
    \centering
    \caption{Hyperparameters for NeuRadar. The hyperparameter values are universal across the three datasets, except for the radar ray divergence. The most suitable value for this hyperparameter is 0.0125 for ZOD and VoD and 0.025 for nuScenes.}
    \vspace{-2mm}
    \resizebox{\linewidth}{!}{
    \begin{tabular}{l l l }
    \toprule
      & Hyperparameter & Value \\ \midrule
    
    \multirow{10}{*}{\rotatebox[origin=c]{90}{Neural feature field}}
    &  RGB upsampling factor  & $3$ \\ 
    &  proposal samples  & $128, 64$ \\
    &  SDF $\beta$  & $20.0$ (learnable) \\
    &  power function $\lambda$    & $-1.0$ \\ 
    &  power function scale  & $0.1$ \\
    &  appearance embedding dim  & $16$ \\
    &  hidden dim (all networks)    &  $32$ \\ 
    &  NFF feature dim  & $32$ \\
    &  Radar ray divergence $\delta_\varphi$ and $\delta_\vartheta$  & $0.0125/0.025$ \\
    &  Radar ray scaling parameter $\zeta$  & $\frac{1}{16}$ \\
    \midrule
    
    \multirow{8}{*}{\rotatebox[origin=c]{90}{Hashgrids}} 
    & hashgrid features per level   &  $4$ \\
    & actor hashgrid levels    & $4$ \\
    & actor hashgrid size & $2^{15}$ \\
    &  static hashgrid levels  & $8$  \\
    &  static hashgrid size   & $2^{22}$ \\
    &  proposal features per level   & $1$ \\
    &  proposal static hashgrid size  &  $2^{20}$ \\
    &  proposal actor hashgrid size   & $2^{15}$ \\
    \midrule

    \multirow{10}{*}{\rotatebox[origin=c]{90}{Loss weights}} 
    & $\lambda^{\text{rgb}}$  &  $5.0$ \\
    & $\lambda^{\text{vgg}}$  &  5e-2 \\
    & $\lambda^{\text{int}}$  &  1e-1 \\
    & $\lambda^{\text{d}}$  &  1e-2 \\
    & $\lambda^{\text{w}}$ &  1e-2 \\
    & $\lambda^{P_d}$ &  1e-2 \\
    & proposal $\lambda^{\text{d}}$ &  1e-3 \\
    & proposal $\lambda^{\text{w}}$  &  1e-3 \\
    & interlevel loss multiplier & 1e-3  \\
    &  $\lambda^{\text{radar}}$ &  2e-2 \\
    \midrule

    \multirow{5}{*}{\rotatebox[origin=c]{90}{Learning rates}} 
    & actor trajectory lr & 1e-3 \\
    & cnn lr   &   1e-3  \\
    & camera optimization lr  & 1e-4 \\
    & transformer lr  & 1e-3 \\
    & remaining parameters lr  &  1e-2 \\

    
    \bottomrule
    \end{tabular}
    }
    \vspace{-2mm}
    \label{tab:hyperparameters}
\end{table}

\begin{table}[ht]
    \centering
    \caption{Results for various maximum offsets in the Cartesian coordinate system. The unit of offset is meters.}
    \resizebox{0.85\linewidth}{!}{
    \begin{tabular}{c c c c c}
    \toprule
     \multirow{2}{*}{Max Offset} & \multicolumn{2}{c}{ZOD} & \multicolumn{2}{c}{VoD} \\  \cmidrule(lr){2-3} \cmidrule(lr){4-5}
      & CD $\downarrow$ & EMD $\downarrow$ & CD $\downarrow$ & EMD $\downarrow$ \\ 
     \midrule
     1.0 & 4.68 & 6.05 & 4.19 & 4.48 \\
     1.5 & \textbf{3.92} & \textbf{4.94} & 4.02 & \textbf{4.28} \\
     2.0 & 4.54 & 6.22 & 4.07 & 4.28 \\
     2.5 & 4.62 & 6.21 & 4.07 & 4.42 \\
     3.0 & 4.55 & 6.14 & \textbf{3.99} & 4.33 \\
    \bottomrule
    \end{tabular}
    }
    \label{tab:ab_study_offset}
\end{table}


\section{Datasets}\label{sec:suppmatdata}

In this section, we provide more detailed information about the datasets used to evaluate our network. 

\subsection{ZOD} \label{sec:zodapp}

We used ZOD sequences \texttt{000030}, \texttt{000546}, and \texttt{000811} for our ablation studies and hyperparameter tuning. For our final experiments we used the following ten sequences: \texttt{000005}, \texttt{000221}, \texttt{000231}, \texttt{000244}, \texttt{000387}, \texttt{000581}, \texttt{000619}, \texttt{000657}, \texttt{000784}, \texttt{001186}. These sequences vary in ego speed, lighting conditions (both day and night), weather conditions (sunny. snowy, and cloudy), and scenario type (highway, city, residential), which we deemed appropriate for our experiments.

The radar point clouds in ZOD are captured every 60 ms and stored in a standard binary file format (.npy) for each ZOD \textit{Sequence} and \textit{Drive}. The data contains timestamps in UTC, radar range in meters, azimuth and elevation angles in radians, range rate in meters per second, amplitude (or SNR), validity, mode, and quality. The radar switches between three modes depending on the ego vehicle speed, and the sensor has a different maximum detection range in each mode. Mode 0 represents the radar point clouds captured when the vehicle speed is less than 60 to 65 kph with a maximum detection range of 102 meters, while modes 1 and 2 represent vehicle speeds of between 60 to 65 kph and 110 to 115 kph, and more than 110 to 115 kph, respectively, with maximum detection ranges of 178.5 and 250 meters. The azimuth angle values are between -50 and 50 degrees. The quality value also changes from 0 to 2, with 2 indicating the highest quality for the detections. The radar extrinsic calibration information (i.e., latitude, longitude, and angle) is provided in calibration files, indicating its position relative to the reference coordinate frame. 

\subsection{VoD}

VoD contains driving scenarios captured at 10 Hz around Delft city from the university campus, the suburbs, and the old town with many pedestrians and cyclists, so the dataset itself is not very diverse and is rather challenging for NeRF-based methods. Since VoD does not provide sequence numbers, we created a set of "sequences", each of which is roughly 30 seconds and originates from a different drive in the dataset. We provide the range of frames numbers in the dataset for each sequence. The VoD sequences used for ablations and hyperparameter tuning were \texttt{1850-2150}, \texttt{7600-7900}, and \texttt{8482-8749}. For our final experiments, we used \texttt{100-400}, \texttt{2220-2520}, \texttt{2532-2798}, \texttt{2900-3200}, \texttt{3277-3575}, \texttt{3650-3950}, \texttt{4050-4350}, \texttt{4387-4652}, \texttt{4660-4960}, and \texttt{6800-7100}. 

\section{Additional Results}\label{sec:resultsapp}

\subsection{Effects on Image and Lidar Rendering}\label{sec:imagelidareffects}

To investigate whether incorporating a radar branch affects image and lidar rendering, we compare the performance of NeuRAD and NeuRadar on image and lidar rendering tasks. The results are shown in \cref{tab:cl_metrics}. We report PSNR, SSIM \cite{1284395}, and LPIPS \cite{Zhang_2018_CVPR} as image similarity metrics for camera simulation. We evaluate the fidelity of lidar simulation using four metrics: L2 median depth error, RMSE intensity error, ray drop accuracy, and Chamfer Distance (CD). For both datasets NeuRadar's performance is similar to NeuRAD and the addition of radar does not affect the camera and lidar reconstruction performance.

\begin{table*}[tb]
    \centering
    \caption{Performance comparison of novel view synthesis for image and lidar. NeuRAD results are obtained using its public code with recommended settings.}
    \vspace{-2mm}
    \resizebox{0.9\linewidth}{!}{
    \begin{tabular}{l c c c c c c c c}
    \toprule
     \multirow{2}{*}{Dataset} & \multirow{2}{*}{Method} & \multicolumn{3}{c}{Camera} & \multicolumn{4}{c}{Lidar} \\  \cmidrule(lr){3-5} \cmidrule(lr){6-9}
      &  & PSNR $\uparrow$ & SSIM $\uparrow$   & LPIPS $\downarrow$ & Depth $\downarrow$ & Intensity $\downarrow$ & Drop acc. $\uparrow$ & Lidar CD $\downarrow$ \\ 
     \midrule
    
    \multirow{4}{*}{ZOD} 
    & NeuRAD    & \textbf{30.9} & \textbf{0.878} & \textbf{0.187} & \textbf{0.028} & \textbf{0.041} & \textbf{95.8} & 3.69 \\ 
    & Baseline     & 29.96 & 0.871 & 0.192 & 0.035 & 0.043 & 95.5 & 3.75 \\ 
    & Deterministic   & 30.4 & 0.870 & 0.190 & 0.035 & 0.045 & 95.6 & 3.73 \\
    & Probabilistic   & 30.1 & 0.870 & 0.191 & 0.030 & \textbf{0.041} & 95.5 & \textbf{3.59} \\
    
    \midrule

    \multirow{4}{*}{VoD} 
    & NeuRAD & 21.68 & \textbf{0.687} & 0.366 & \textbf{0.167} & \textbf{0.158} & 85.78 & 13.96 \\ 
    & Baseline   & 21.59 & 0.680 & 0.372 & 0.217 & 0.158 & 85.73 & 14.45 \\
    & Deterministic   & 21.63 & 0.683 & 0.365 & 0.172 & 0.163 & \textbf{85.86} & 11.92 \\
    & Probabilistic   &\textbf{ 21.75} & 0.683 & \textbf{0.363} & \textbf{0.167} & 0.159 & 85.75 & \textbf{11.66} \\
    
    \bottomrule
    \end{tabular}
    }
    \label{tab:cl_metrics}
\end{table*}

\subsection{DETR as Radar Decoder}\label{sec:detrdecoder}

\begin{figure}
    \centering
    \includegraphics[width=0.8\linewidth]{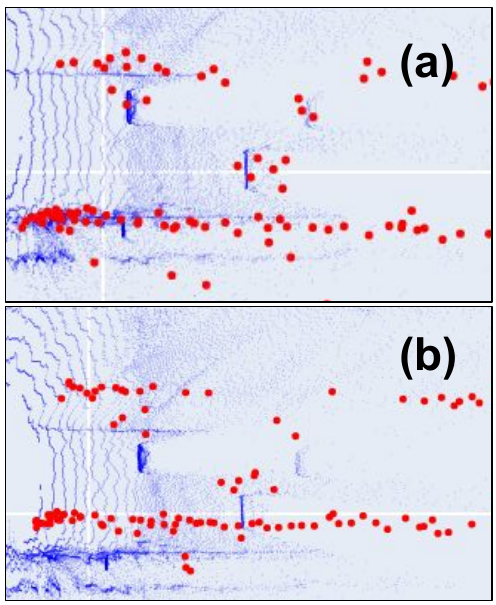}
    \caption{Radar point cloud (in red) rendered by the naive DETR-based radar decoder depicted along rendered lidar point cloud. (a) shows the output for an interpolated sensor pose in a ZOD sequence, and (b) shows the radar detections rendered with a 2 meter ego pose shift. The radar detections have merely shifted in position and do not reflect the geometry of the scene as shown by lidar.}
    \label{fig:detr_issues}
\end{figure}

As mentioned in \cref{sec:radarmodeling}, a potential solution to radar modeling is to directly predict radar detections, or in this case MB parameters, from the NFF features using a network commonly used for object detection. To further explore this idea, we evaluate novel view synthesis using a DETR-like transformer network \cite{carion2020endtoend} as the radar decoder, where the number of output queries is equal to the maximum potential number of radar detections. In \cref{fig:detr_issues} we show the radar and lidar detections generated by this network. Although the method can predict a reasonable point cloud from previously seen viewpoints, it is completely incapable of true novel view synthesis, implying that the network ignores the geometric info about the scene and has merely learned to copy the ground truth regardless of features. 

\subsection{Visualizations}\label{sec:visualizations}

\begin{figure*}
    \centering
    \includegraphics[width=0.9\linewidth]{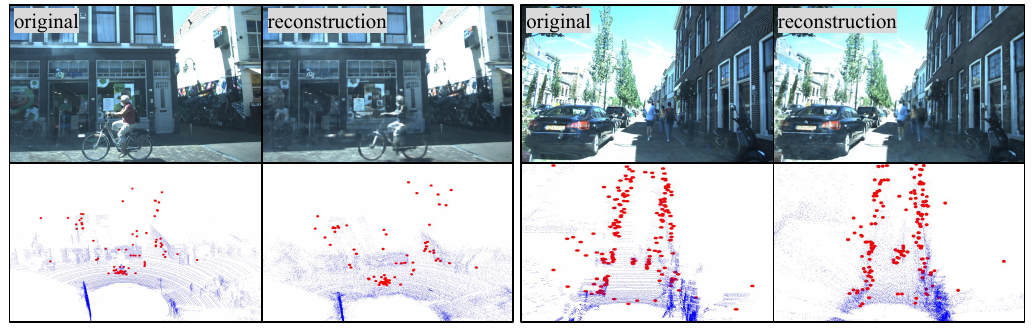}
    \caption{Novel view synthesis results VoD sequences.}
    \label{fig:vod_vis}
\end{figure*}

In this section, we visualize the qualitative results of our experiments. \cref{fig:vod_vis} shows the novel view synthesis results for two VOD sequences using our probabilistic method. VOD is a dataset specifically curated for urban scenarios with many pedestrians and cyclists, making it a challenging dataset to use for NeRF-based methods. 

\section{Limitations and Future Work}\label{sec:limits}

In this work, NeuRadar effectively generates realistic radar point clouds. However, certain characteristics of radar data are not fully captured. Here, we describe two limitations of our work, which also point to directions for enhancing NeuRadar.

First, NeuRadar's radar decoder does not predict radar range rate, signal-to-noise ratio (SNR), or radar cross section (RCS). We believe that extending NeuRadar to incorporate these values and addressing the associated challenges could make a valuable contribution. Another limitation stems from the inherent mechanisms of NeRFs. Designed primarily for visual data, NeRFs focus on surface geometry and visible features, which hinders NeuRadar's ability to fully leverage radar's strength in detecting visually occluded objects. Addressing this limitation presents a promising research direction.